# Predicting Li-ion Battery Cycle Life with LSTM RNN

Pengcheng Xu, Yunfeng Lu[*]
Department of Chemical and Biomolecular Engineering, University of California, Los Angeles, 90095
{phoenixfilber, luucla}@ucla.edu

## 1. Abstract

Efficient and accurate remaining useful life prediction is a key factor for reliable and safe usage of lithium-ion batteries. This work trains a long short-term memory recurrent neural network model to learn from sequential data of discharge capacities at various cycles and voltages and to work as a cycle life predictor for battery cells cycled under different conditions. Using experimental data of first 60 - 80 cycles, our model can achieve promising prediction accuracy on test sets of around 80 samples.

## 2. Introduction

Lithium-ion batteries are widely used in electric devices because of their high energy densities, low cost and long lifetime [1]. As an important functionality of management systems of batteries, remaining useful life prediction gives probable failure time in advance for diagnostics and prognostics and aids manufacturing and operation of battery cells and systems. Meanwhile, it is also a challenging task since capacity degradation of batteries is a complex and nonlinear process influenced by internal physics and operating conditions.

There are many excellent research works for prediction of remaining useful life of batteries, where model-based methods and data-driven methods are two major branches. Model-based methods builds up mathematical models or semi-empirical models to capture the relations among internal processes, operation conditions and battery capacity degradation. First principles-based aging models are developed in [2][3][4], where relations between kinetics of SEI layer formation, reduction of electrolyte volume fraction and loss of cyclable lithium are explored. Particle filter [5] is a widely used fundamental algorithm combining with other theories and techniques for further improvement such as Dempster Shafer theory [6] and Akaike information criterion [7]. However, model-based methods are still restricted by lack of accurate aging models and particle degeneracy problem.

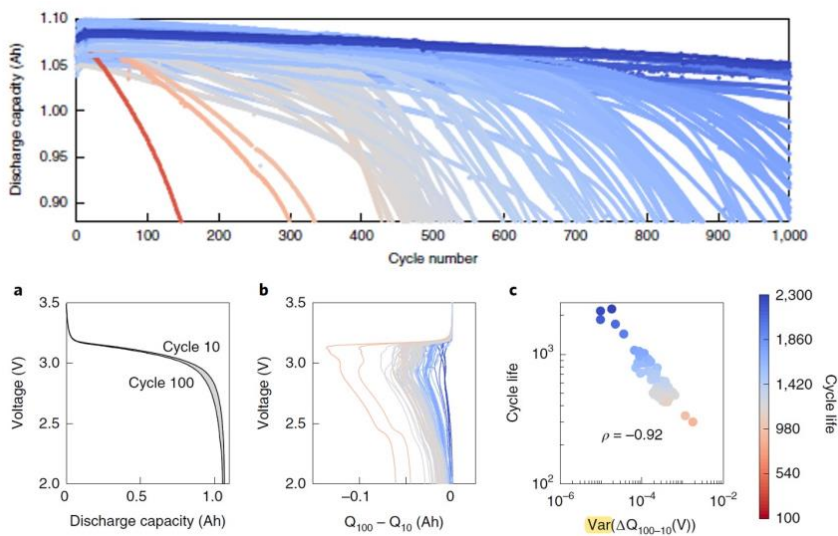

Figure 1 Visulation of the data set provided by literature. [8]

Data-driven methods, compared to model-based methods, do not require a mathematical or semi-empirical model but depend on experimental data of battery cycling. It is of essential significance to extract to relevant features from available data for life prediction. Severson and coworkers [8] explore a new feature about variations of discharge capacities as functions of cycles and voltages, which has a high correlation coefficient with cycle life. Combined with classical linear regression method, it gives promising results of cycle life prediction before obvious capacity degradation by only using data of first 100 cycles. The feature discussed in [8] only considers the difference between cycle 100 and cycle 10. It is interesting to see whether there will be any improvement on model prediction results (e.g, use even fewer cycles to achieve higher accuracy) by incorporating more cycles (cycle 20, cycle 30 and so on). In this way, the discharge capacities of various voltages under different cycles could be depicted as a sequence: a time step is a cycle, and the discharge capacities (of various voltages) for this cycle are features of this time step. Thus, the problem of interest is to use a sequence of battery discharge capacities (as functions of voltage and cycle) to predict the corresponding cycle life.

To learn from sequential data of capacity degradation, a suitable model plays an important role. recurrent neural networks (RNN) could be a probable option since there is an internal state capable to represent information of capacity degradation, which has been discussed by [9], [10]. Furthermore, long short-term memory (LSTM) RNN, with a solution to the problem of "gradient vanishing", is a powerful technique for learning long-term dependencies (how capacities degradation gets developed thorough cycles). Zhang and coworkers [11] present an application of LSTM RNN for battery remaining useful life predictions, which outperforms the results of support vector machine (SVM) and traditional RNNs.

This work tries to combine some significant characteristics of these excellent research works by feeding the sequential data of discharge capacities at various voltages and cycles to a LSTM RNN model, which exhibits the promise of achieving following improvements:

- capture latent information from discharge capacity vs voltage curves of early cycles;
- predict cycle lives of acceptable accuracy with fewer required input;
- reduce the need to manually extract task-specific features.

## 3. Battery Data Set

The data set used in this work is provided by [8]. As the largest public battery dataset, it contains 124 identical commercial lithium iron phosphate/graphite cells (manufactured by A123 Systems) cycled under 72 different fast-charging and identical discharging conditions. The split of training set and test set is consistent with the original paper: 41 samples in the training set, 43 samples in the primary test set, and 40 samples in the secondary test set.

As Figure 2 shows, the input of the LSTM RNN model is a sequence of discharge capacities at different voltages. For each sample of battery cell, data from a starting cycle (e.g., cycle 11) to a terminal cycle (e.g., cycle 100) are used as time steps. In each time step, discharge capacities at voltages within a range of [3.5 V, 2.0 V] (with 0.01 V as step size) are chosen as features.

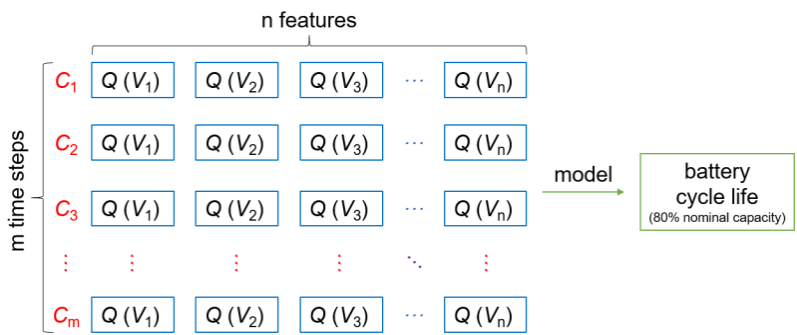

Figure 2 LSTM RNN model input (left) and output (right): For input, Q is discharge capacity, which is a function of voltage V and cycle number C. A series of Cs (monotonic increasing, not necessarily consecutive) are viewed as time steps, and Qs (of Vs) given a specific C are viewed as features of this time step. For output, under cycling of assigned charging and discharging, the cycle at which a battery cell remains 80% of its nominal capacity is defined as its cycle life for the given cycling condition.

At discharge stages of various cycles, the discharge capacity under a specific voltage also varies due to capacity degradation. Intuitively, as Figure 3 shows, within a range of cycles, battery cells showing longer cycle lives have more "uniform" discharge capacity vs voltage curves while those with shorter cycle lives have more "dispersed" discharge capacity vs voltage curves. In addition to normalizing data of discharge capacities for the ease of neural network learning, in order to make the differences of discharge capacities among difference cycles more obvious, we use the discharge capacities at cycle 10 as base values and subtract them from those at other cycles.)

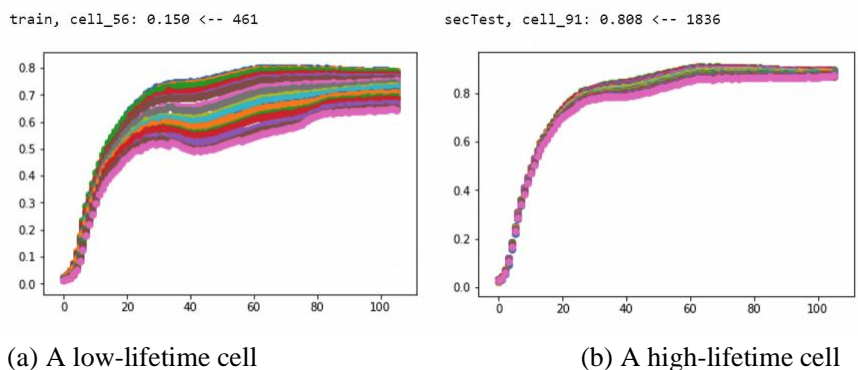

(a) A low-lifetime cell　　　　　　　　　　　(b) A high-lifetime cell

Figure 3 Examples of discharge capacity vs voltage curves for battery cells with different lifetimes: A curve stands for a cycle (the higher, the earlier cycle); x-axis is normalized voltage, y-axis is normalized discharge capacity.

## 4. LSTM RNN MODEL

**Architecture**

In our LSTM RNN model built by Keras with TensorFlow backend [12], an input layer is fed into an CuDNNLSTM layer with 64 - 128 neurons, then fed into another CuDNNLSTM layer with 128 - 256 neurons, next fed into a densely-connected layer of 32 neurons with a "relu" (Rectified Linear Unit) activation function, and finally fed into a densely-connected layer of 1 neuron with a linear activation function (since the task here is prediction instead of classification). It is worth mentioning that CuDNNLSTM layers are employed here instead of normal LSTM layers for faster model training, prediction and prediction [13].
A general issue in machine learning application is to avoid overfitting. Since there are small data set and large number of model parameters, it is easy for the model to become overfitted. Dropout technique [14] is employed to prevent overfitting, and the retained probability of hidden layers of was set to 0.2.

**Training**

In our neural network model, the loss function is mean squared error (MSE). It is defined as

$$\text{MSE} = \frac{1}{N} \sum_{i=1}^{N} (\hat{S}_i - S_i)^2$$

where $N$ is the example number of test set, $S$ is the set of predicted cycles lives while $\hat{S}$ is the set of actual cycle lives.
Adam optimizer [15] is used to train the model (learning rate is 1e-3, and decay rate is 1e-6). For gradient descent, the batch size is set to 128, and number of epochs is 400 - 1000.
The training was implemented on a HP Z420 workstation with an Intel Xeon E5-1650 v2 processor (up to 3.50 GHz) and a graphic card of Nvidia GTX 1080 Ti (11 GB memory).
In order to balance low-lifetime samples with high-lifetime ones and prevent overfitting of the training set, a strategy of data augmentation is applied to generate "fake" samples by "shifting" genuine samples. For example, given data of battery cell, a genuine sample takes cycle 11 - cycle 60 as model input and its cycle life (denoted as cl) as model output, and then a fake sample can use cycle 16 - cycle 65 as model input and (cl - 5) as model output, which is obtained "shifts" the genuine sample by 5 cycles. The step of "shifting" is 3 cycles, if applied.

## 5. Results and Discussion

Figure 4, Table 1 and Table 2 evaluates our model (taking inputs with different terminal cycles from 40 to 100) on 2 kinds of evaluation metrics, compared with classical models developed in [8]. The applied evaluation metrics consists of root mean squared error (RMSE) and mean

absolute percentage error (MAPE) for our LSTM RNN model. RMSE is defined as the square root of MSE, and MAPE is defined as

$$\text{MAPE} = \frac{1}{N}\sum_{i=1}^{N}(\hat{S}_i - S_i)^2/\hat{S}_i \times 100\%$$

where $N$, $S$ and $\hat{S}$ have same meanings to previous formula. For referenced models, "variance" model utilizes only 1 feature having high correlation coefficient with cycle lives and achieves good prediction results. The feature is

$$\log \text{var}\,(Q_{100}(V) - Q_{10}(V))$$

where $Q$ is discharge capacity, $V$ is voltage, 100 and 10 are cycle numbers. Moreover, "discharge" model adds more features extracted from current density and voltage during discharging stages in the first 100 cycles, and it presents even better ability to predict cycle lives. In our LSTM RNN model, for both test sets and both evaluation metrics, there seems to be a general tendency that the errors become relatively flat after cycle 60 - 80. For the primary test set, our model shows good results (between 2referenced models) by using data of first 60 cycles (terminal cycle is 60), and it gets similar results to "discharge" model when using data of first 80 cycles. It seems that, for the primary data set, our model is capable to give good predicted cycle lives using data of fewer than 100 cycles.

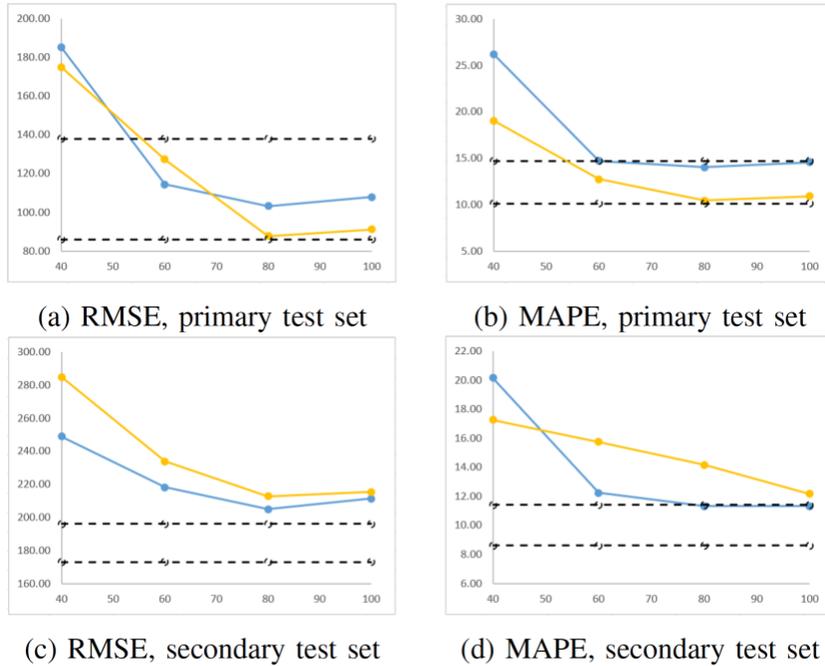

(a) RMSE, primary test set  (b) MAPE, primary test set

(c) RMSE, secondary test set  (d) MAPE, secondary test set

Figure 4 Model results: x-axis is terminal cycle, RMSE is root mean square error, and MAPE is mean average percentage error; black dotted lines are referenced results from [8]: upper line stands for "variance" model, and lower line stands for "discharge" model; colored lines are results of our LSTM RNN model (dots are average values over 10 parallel experiments): yellow line takes non-augmented data for training, blue line takes augmented data.

For the secondary test set, however, things become interesting. Using same training set consisting of 41 samples, our model does not give satisfactory prediction results, which are even not close to those of "variance" model. A possible explanation is that, the secondary test set contains some "latent" features quite different from the training set. Due to the small amount of available samples and large number of trainable parameters in our model, the trained parameters overfits the training set and cannot generate well to the secondary test set.

Table 1 Model Result Comparison on RMSE (Cycle).

| Model | Train | PriTest | SecTest |
|---|---|---|---|
| "Variance" (@ 100) [8] | 103 | 138 (138) | 196 |
| "Discharge" (@ 100) [8] | 76 | 91 (86) | 173 |
| LSTM RNN (@ 60) | 51.1 ±0.1 | 127.3 ±1.4 | 223.8 ±0.5 |
| LSTM RNN (@ 80) | 47.9 ±2.1 | 87.7 ±6.0 | 212.8 ±2.2 |
| LSTM RNN (@ 100) | 42.3 ±1.0 | 91.2 ±3.8 | 215.3 ±3.5 |
| LSTM RNN (@ 60, aug) | 63.4 ±5.0 | 114.3 ±3.5 | 218.2 ±4.0 |
| LSTM RNN (@ 80, aug) | 63.3 ±2.8 | 103.1 ±2.7 | 204.9 ±2.7 |
| LSTM RNN (@ 100, aug) | 56.3 ±6.3 | 107.9 ±5.3 | 211.3 ±5.6 |

Note: "@" is followed by terminal cycle of input data, and "aug" means augmented data are used during model training. Same note to the other table.

Table 2 Model Result Comparison on MAPE (%).

| Model | Train | PriTest | SecTest |
|---|---|---|---|
| "Variance" (@ 100) [8] | 14.1 | 14.7 (13.2) | 11.4 |
| "Discharge" (@ 100) [8] | 9.8 | 13.0 (10.1) | 8.6 |
| LSTM RNN (@ 60) | 6.2 ±0.1 | 12.7 ±0.4 | 15.8 ±0.1 |
| LSTM RNN (@ 80) | 6.2 ±0.3 | 10.4 ±0.4 | 14.2 ±0.7 |
| LSTM RNN (@ 100) | 5.6 ±0.2 | 10.9 ±0.7 | 12.2 ±0.2 |
| LSTM RNN (@ 60, aug) | 7.9 ±1.0 | 14.7 ±0.9 | 12.2 ±0.2 |
| LSTM RNN (@ 80, aug) | 8.7 ±0.6 | 14.0 ±0.5 | 11.3 ±0.6 |
| LSTM RNN (@ 100, aug) | 7.1 ±1.0 | 14.6 ±1.2 | 11.3 ±0.4 |

A preliminary trial of data augmentation is applied to reduce the phenomena of overfitting. Since there is less high-lifetime samples (having cycle life greater than 775, which is just a causal trial) in the training set than in the secondary test set, the trick to augment data (mentioned in Section III) is applied to high-lifetime samples but not low-lifetime ones. Our model trained with augmented training set shows improved prediction results on the secondary test set: it approaches "variance" model by only using the first 80 cycles of data, but there is still a gap to "discharge" model. Unexpectedly, our model trained with augmented data gives worse results on the primary data set, which may results from the "enlarged" dissimilarity between the training set and the primary test set after augmenting the training set.

Admittedly, our model has far more complex structure and consumes more computation resources for training than classical models presented by [8]. However, it has the potential to learn from complex sequential data of battery cycling so that it might require fewer number of input cycles to make predictions of similar accuracy compared to classical models, where those "saved" cycles could mean a lot in industrial research and development. Moreover, unlike "discharge" model, out model does not intake additional "task-specific" features such as skewness and kurtosis of cycling data of discharge capacity vs voltage curves. Although our model currently cannot give as excellent results as classical models on the secondary test set, with more available data and advanced overfitting-preventing techniques, its ability to generalize can be further improved.

## 6. Conclusion

In this work, an LSTM RNN model is developed for early prediction of battery cycle lives. The input data is cycle sequences of capacities within discharge voltage windows, which is inspired by the feature highly correlated with cycle lives presented by [8]. By taking first 60 - 80 cycles of data, our model achieves satisfactory RMSE and MAPE on the primary test set. With data augmentation during training, the model also shows acceptable prediction results on the secondary test set by using fewer cycles of data. In general, this work explores a LSTM RNN model taking a new input form of sequential data, which is promising in reducing the required number of early cycles for predicting cycle lives.